\documentclass[11pt,a4paper]{article}
\usepackage{acl2020}
\usepackage{times}
\usepackage{graphicx}
\usepackage{wrapfig}
\usepackage{shortcuts}
\usepackage{dsfont}
\usepackage{latexsym}
\usepackage[linesnumbered,algoruled,boxed,noend]{algorithm2e}
\usepackage{amssymb}
\usepackage{amsmath}
\usepackage{url} 
\SetKwInOut{Parameter}{parameter}
\usepackage{enumitem}
\usepackage{mathtools}
\usepackage{cleveref}
\usepackage{multirow}
\usepackage{hhline}
\usepackage[export]{adjustbox}
\usepackage{booktabs,array}
\usepackage{amsmath, bm}
\usepackage{color, colortbl}
\usepackage{subfig}
\SetKwInput{KwInput}{Input}
\SetKwInput{KwOutput}{Output}
\newcolumntype{P}[1]{>{\centering\arraybackslash}p{#1}}
\usepackage{booktabs,makecell,tabularx} 
\setitemize{noitemsep,topsep=0pt,parsep=0pt,partopsep=0pt}
\let\oldnl\nl
\definecolor{Gray}{gray}{0.85}
\definecolor{LightCyan}{rgb}{0.88,1,1}
\definecolor{FigCsqaOrange}{RGB}{236, 215, 192}
\definecolor{FigRsBlue}{RGB}{191, 207, 255}
\newcommand{\nonl}{\renewcommand{\nl}{\let\nl\oldnl}}

\usepackage{pifont}
\aclfinalcopy 
\def\aclpaperid{1126}
\usepackage{epigraph}

\usepackage{etoolbox}
\patchcmd{\epigraph}{\@epitext{#1}}{\itshape\@epitext{#1}}{}{}

\newlength{\Width}%
\newlength{\DepthReference}
\newlength{\HeightReference}
\newcommand{\MyColorBox}[2][red]%
{%
    \settowidth{\Width}{#2}%
    \colorbox{#1}%
    {%
        \raisebox{-\DepthReference}%
        {%
                \parbox[b][\HeightReference+\DepthReference][c]{\Width}{\centering#2}%
        }%
    }%
}
\begin{document}

\title{\vspace*{-0.5in}
{{\small \hfill ACL-IJCNLP 2021 Findings}\\
\vspace*{.25in}}
RiddleSense: Reasoning about Riddle Questions \\ Featuring Linguistic Creativity and Commonsense Knowledge}

\author{
Bill Yuchen Lin \quad Ziyi Wu\quad Yichi Yang\quad  Dong-Ho Lee\quad Xiang Ren\\
\texttt{\{yuchen.lin,ziyiwu,yichiyan,dongho.lee,xiangren\}@usc.edu}\\
Department of Computer Science and Information Sciences Institute,  \\ University of Southern California\\
}


\maketitle

\begin{abstract}

Question: \textit{I have five fingers but I am not alive.  What am I?}  Answer: \textit{a glove}. 

Answering such a riddle-style question is a challenging cognitive process, in that it requires complex commonsense reasoning abilities, an understanding of figurative language, and counterfactual reasoning skills, which are all important abilities for advanced natural language understanding (NLU).
However, there is currently no dataset aiming to test these abilities.
In this paper, we present \textsc{RiddleSense}\footnote{\url{https://inklab.usc.edu/RiddleSense/}}, a new multiple-choice question answering task,
which comes with the first large dataset (5.7k examples) for answering riddle-style commonsense questions. 
We systematically evaluate a wide range of  models over the \textsc{RiddleSense} challenge, and point out that there is a large gap between the best-supervised model and human performance --- suggesting intriguing future research in the direction of higher-order commonsense reasoning and linguistic creativity towards building advanced NLU systems. 



\end{abstract}

\section{Introduction}\label{sec:intro}
\epigraph{\normalsize ``\textit{ \textbf{The essence of a riddle is to express true facts under impossible combinations.}}"}{\normalsize--- \textit{Aristotle}, \textit{Poetics} (350 BCE)\vspace{0pt}}

\noindent
A \textit{riddle} is a puzzling question about {concepts} in our everyday life.
For example, a riddle might ask ``\textit{My life can be measured in hours. I serve by being devoured. Thin, I am quick. Fat, I am slow. Wind is my foe. What am I?}''~
The correct answer ``\textit{candle},'' is reached by considering a collection of \textit{commonsense knowledge}:
{a candle can be lit and burns for a few hours; a candle's life depends upon its diameter; wind can extinguish candles, etc.}
\begin{figure}[t]
	\centering 
	\includegraphics[width=1\linewidth]{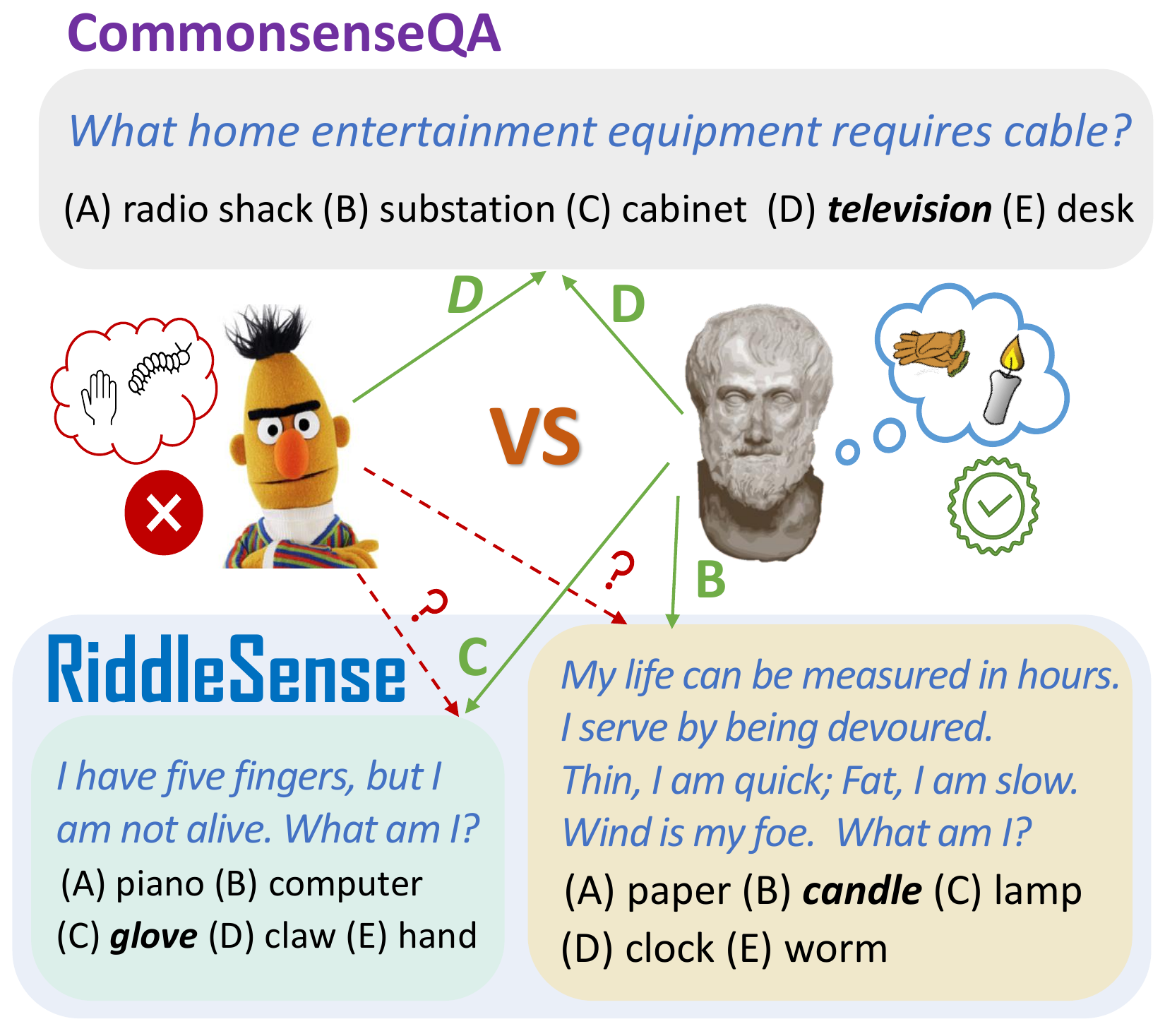}
	\caption{ 
    The top example is a trivial commonsense question from the CommonsenseQA~\cite{Talmor2018CommonsenseQAAQ} dataset. 
    The two bottom examples are from our proposed \textsc{RiddleSense} challenge.
    The right-bottom question is a descriptive riddle that implies multiple commonsense facts about \textit{candle}, and it needs understanding of figurative language such as metaphor;
    The left-bottom one additionally needs counterfactual reasoning ability to address the \textit{`but-no'} cues. 
    These riddle-style commonsense questions  require NLU systems to have higher-order reasoning skills with the understanding of creative language use.
	}
	\label{fig:intro} 
\end{figure}

It is believed that the \textit{riddle} is one of the earliest forms of oral literature,
which can be seen as a formulation of thoughts about common sense, a mode of association between everyday concepts, and a metaphor as higher-order use of natural language~\cite{hirsch2014poet}.
Aristotle stated in his \textit{Rhetoric} (335-330 BCE) that good riddles generally provide satisfactory metaphors for rethinking common concepts in our daily life.
He also pointed out in the \textit{Poetics} (350 BCE): ``the essence of a riddle is to express true facts under impossible combinations,'' which suggests that solving riddles is a nontrivial  reasoning task.

Answering riddles is indeed a challenging cognitive process as it requires \textit{complex} {commonsense reasoning skills}.
A riddle can describe \textit{multiple pieces} of commonsense knowledge with \textit{figurative} devices such as metaphor and personification (e.g., ``wind is my \underline{foe} $\xrightarrow[]{}$ \textit{extinguish}'').
Moreover, \textit{counterfactual thinking} is also necessary for answering many riddles such as ``\textit{what can you hold in your left hand \underline{but not} in your right hand? $\xrightarrow[]{}$ your right elbow.}''
These riddles with \textit{`but-no'} cues require that models use counterfactual reasoning ability to consider possible solutions for situations or objects that are seemingly impossible at face value.
This \textit{reporting bias}~\cite{gordon2013reporting} makes riddles a more difficult type of commonsense question for pretrained language models to learn and reason.
In contrast, \textit{superficial} commonsense questions such as ``\textit{What home entertainment equipment requires cable?}'' in  CommonsenseQA~\cite{Talmor2018CommonsenseQAAQ} are more straightforward and explicitly stated.
We illustrate this comparison in Figure~\ref{fig:intro}.

In this paper,
we introduce the \textsc{RiddleSense} challenge 
to study the task of answering riddle-style commonsense questions\footnote{We use ``riddle'' and ``riddle-style commonsense question'' interchangeably in this paper.} requiring \textit{creativity}, \textit{counterfactual thinking} and \textit{complex commonsense reasoning}.
\textsc{RiddleSense} is presented as a \textit{multiple-choice question answering} task where a model selects one of five answer choices to a given riddle question as its predicted answer, as shown in Fig.~\ref{fig:intro}.
We construct the dataset by first crawling from several free websites featuring large collections of human-written riddles and then aggregating, verifying, and correcting these examples using a combination of human rating and NLP tools to create a dataset consisting of 5.7k high-quality examples.
Finally, we use \textit{Amazon Mechanical Turk} to crowdsource quality distractors to create a challenging benchmark.
We show that our riddle questions are more challenging than {CommonsenseQA} by analyzing graph-based statistics over ConceptNet~\cite{Speer2017ConceptNet5A}, a large knowledge graph for common sense reasoning.


Recent studies have demonstrated that
 fine-tuning large pretrained language models, such as {BERT}~\cite{Devlin2019}, RoBERTa, and ALBERT~\cite{Lan2020ALBERT}, can achieve strong results on current commonsense reasoning benchmarks.
Developed on top of these language models, graph-based language reasoning models such as KagNet~\cite{kagnet-emnlp19} and MHGRN~\cite{feng2020scalable} show superior performance. 
Most recently, UnifiedQA~\cite{khashabi2020unifiedqa} proposes to unify different QA tasks and train a text-to-text model for learning from all of them, which achieves state-of-the-art performance on many commonsense benchmarks.

To provide a comprehensive benchmarking analysis, we systematically compare the above methods.
Our experiments reveal that while humans achieve 91.33\% accuracy on \textsc{riddlesense}, the best language models can only achieve 68.80\% accuracy, suggesting that there is still much room for improvement in the field of solutions to complex commonsense reasoning questions with language models.
We believe the proposed \textsc{RiddleSense} challenge suggests productive future directions for machine commonsense reasoning as well as the understanding of higher-order and creative use of natural language.

\section{Construction of \textsc{RiddleSense}}
\label{sec:datagen}
In this section,
we first present our pipeline for collecting the \textsc{RiddleSense} dataset, including the details of data cleaning.
We introduce how we design a crowd-sourcing protocol for annotating quality distractors to turn riddle-solving into a multiple-choice question answering task.

\subsection{Riddle Crawling and Cleaning}

We write web crawlers for collecting a large number (approximately 10,000) of riddles and their answers from public riddle websites, such as \textit{brainzilla.com}, \textit{riddlewot.com}, etc.
As the crawled data contain much noise such as inconsistent answer format and misspelled words, we process riddles through careful data cleaning as well as human verification.
First, we use an open-source tool for detecting typos\footnote{\url{github.com/phatpiglet/autocorrect}} and then refine the sentences.
Then we continuously sample (riddle, answer) pairs and recognize errors, for which we iteratively improve our program with a set of conditions to filter out noisy examples that are not readable or have ambiguous answers.
Also, we merge the riddles from different sources while removing duplicate riddle questions with similar answers.
For detecting duplicate riddles with minor word changes, we use SentenceBERT~\cite{reimers2019sentence} to find clusters with high cosine similarities.


\subsection{Distractor Collection from AMT}
We consider a multi-choice question answering format rather than the open-ended format, as it is easier to meaningfully compare the performance of different models in a more controlled manner --- there is a limited range of of options. For such a dataset, given a riddle-style question and 5 answer options, the model should select the best one as the predicted answer. 
This format offers a straightforward and fair evaluation metric -- \textit{accuracy}, which is the metric adopted by  many popular commonsense reasoning benchmarks such as CommonsenseQA, ARC~\cite{Clark2018ThinkYH}, and OpenbookQA~\cite{Mihaylov2018CanAS}. 

High-quality distractors are essential for multiple-choice question answering tasks as they can ensure that the dataset is both \textit{clean} and \textit{challenging} --- the distractors are neither too similar nor too distant from the correct answer.
We thus design a protocol to collect quality distractors from human annotators via \textit{Amazon Mechanical Turk}\footnote{\url{https://www.mturk.com/}} based on a pool of candidate distractors.

\paragraph{Candidate Distractor Pool}
We use $Q$ to denote the concepts that are mentioned in the question, and $a$ to denote the concept in the answer\footnote{If there are multiple concepts, we pick the one with the least network degrees 
as they tend to be more important.}.
We then first get all two-hop neighbors in the ConceptNet of $a$ and one-hop neighbors of each $c \in Q$ respectively:
\begin{align*}
    A &= \{x | (x, r_i, y), (y, r_j, a) \} \\
    B &= \{x | (x, r_k, c), \forall c\in Q \} \\
   &~~~~D = A \cap B ,
\end{align*}
where $r_{i/j/k}$ is a binary relation in the ConceptNet such as \texttt{HasProperty}.
The final intersection, $D$, is thus the pool of distractor candidates.
We further use \textit{WordNet}~\cite{miller1995wordnet} to filter out concepts that have either too low or too high \textit{Wu-Palmer} similarity\footnote{We use 0.5 as a threshold which is effective as expected.}.
We argue that such sampled distractors are semantically relevant to both questions and answers, and are also closer to answers in the WordNet taxonomy.
Thus, they are more likely to serve as ideal distractors in a multiple-choice question answering task. 

\paragraph{AMT Crowd-sourcing}
We design a three-stage annotation protocol:

\begin{itemize} 
    \item S1) {\textbf{Sanity Check}}. We show a question and 3 choices where only 1 choice is correct and the other 2 are randomly sampled concepts from the full vocabulary of ConceptNet. 
    Only when the workers pass this sanity check, their following annotations will be considered, so we 
    can avoid noise from random workers.
    
    \item S2) \textbf{Candidate Selection}. As it is difficult to control and verify the quality of distractors from crowd workers, 
we first sample concepts from ConceptNet, which are relevant to both question concepts and answer concepts, forming a set of candidate distractors $D$ for annotators to choose from.
Workers are required to select at least 5 concepts that they think are good distractors to the question. There are at least 3 different workers for each question and we take the candidates which are selected by at least two different workers to make sure the selected distractors are indeed meaningful.
    
    \item S3) \textbf{Open Distractor Collection}. 
    We also ask \textit{master workers} on AMT to write at least one more distractor based on the question context.
    This stage is important because sometimes the candidate pool contains fewer candidates of good quality and the human-written distractors are usually better than the ones in the candidate pool.
    We thus give extra bonus credits to encourage annotators to write more quality distractors.
\end{itemize}

\begin{table}[t]
\centering
	\scalebox{1
	}{
\begin{tabular}{@{}r|cc}
\toprule
                                            & \textbf{CSQA}      & \textbf{RS}       \\ \midrule
\# All    Examples                                  & 12,102     & 5,715      \\
\# Train Examples                                   & 9,741      & 3,510      \\
\# Validation  Examples                                    & 1,221      & 1,021      \\
\# Test  Examples                                   & 1,140      & 1,184      \\ \midrule
Average Question Length                     & 15.06     & 24.04 \\ 
\% Long Qs ($>$20 tokens) & 16.5\%   & 47.3\%   \\

Distinct Question Words                     & 6,822      & 7,110      \\
Distinct Choice Words                       & 7,044      & 9,912    \\
Avg PLL of Qs & -34.41 & -53.98 \\
\midrule
QA-NLI Conflict & 12.7\% & 39.6\% \\
QA-NLI Neutral & 71.6\% & 44.9\% \\
QA-NLI Entailment & 15.7\% & 15.5\% \\
 \bottomrule
\end{tabular}
}
\caption{Key statistics of the \textsc{RiddleSense} dataset (v1.1) vs the CommonsenseQA (CSQA) dataset.}
\label{tab:stat}
\end{table}

\begin{figure*}[th]
	\centering
    \subfloat[\centering Length of Q-A Paths]{{\includegraphics[width=3.75cm]{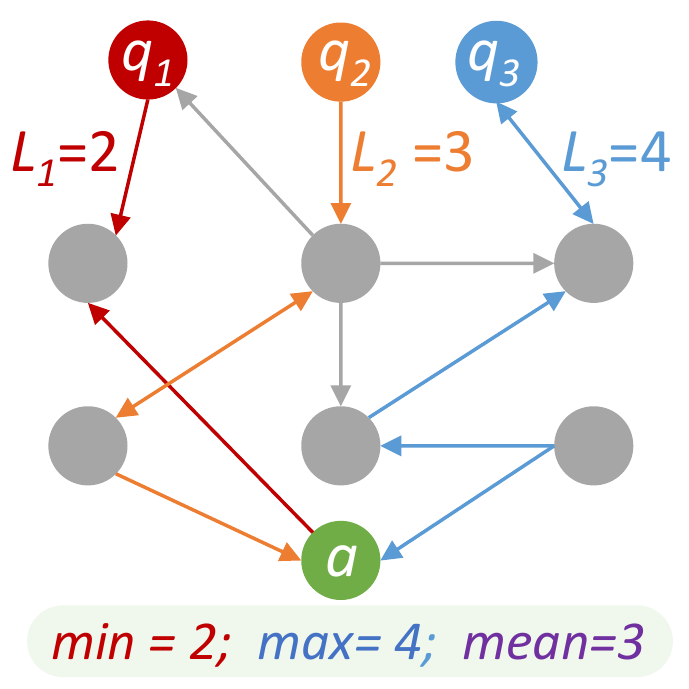} }}
    \subfloat[\centering Mean Length]{{\includegraphics[width=3.9cm]{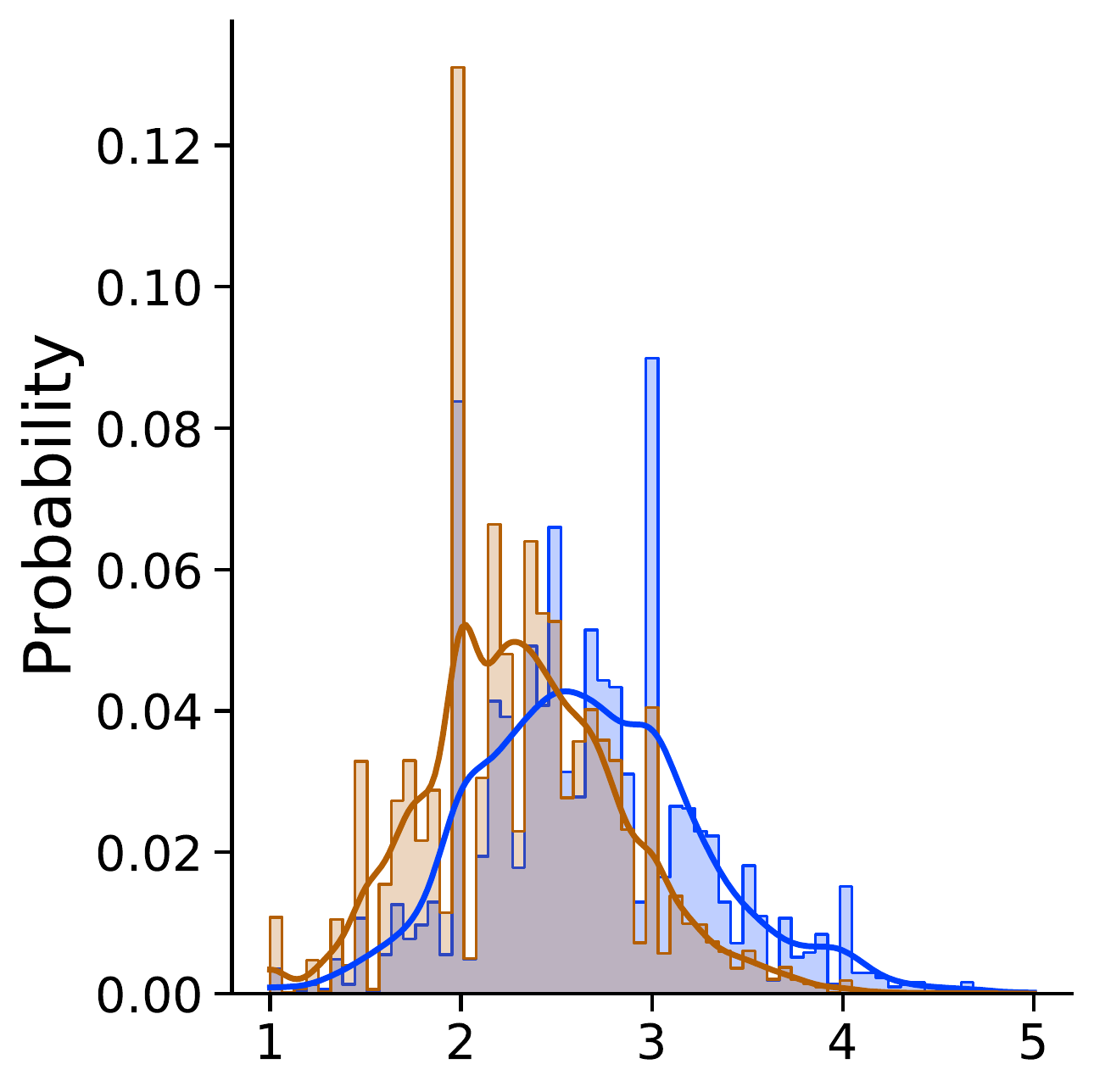} }}
    \subfloat[\centering Min Length]{{\includegraphics[width=3.9cm]{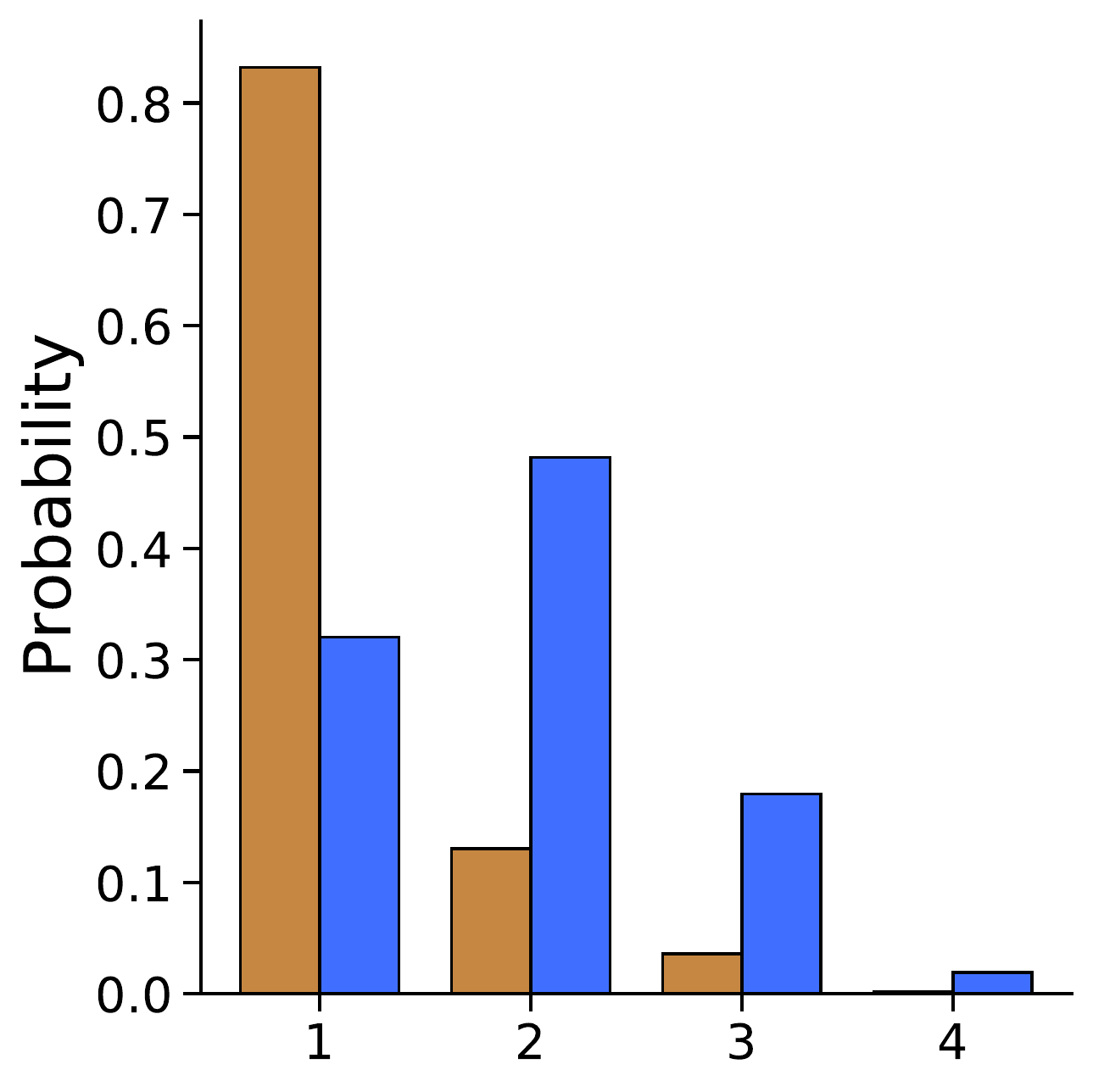} }}
    \subfloat[\centering Max Length]{{\includegraphics[width=3.9cm]{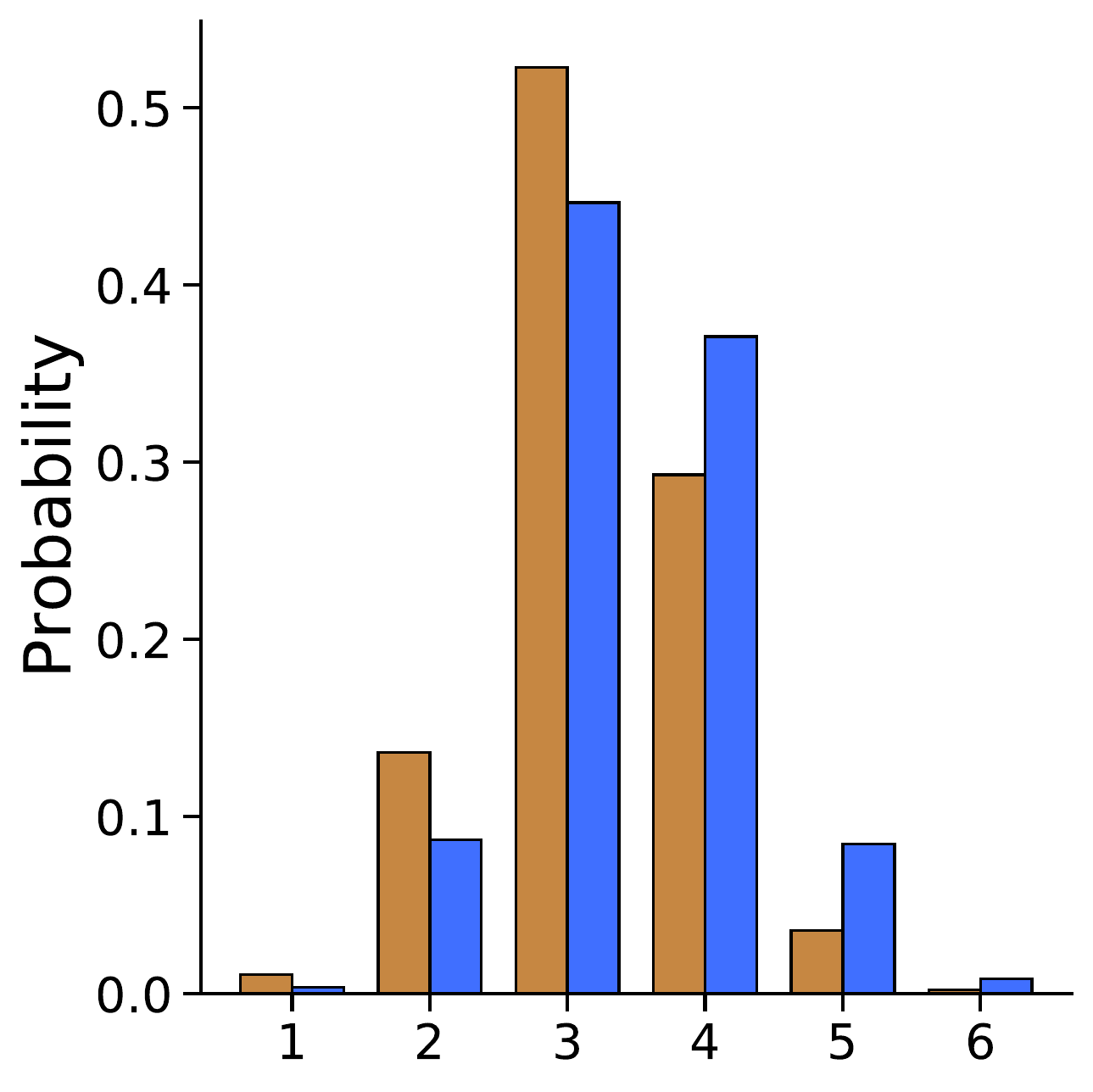} }}
	\caption{The Q-A paths serve as estimation of underlying reasoning chains. Fig.\ (a) illustrates how to compute mean/min/max of the Q-A paths: $\{q_1,q_2,q_3\}$ are three concepts mentioned in the question, and $a$ is the answer concept. $L_k$ is the length of the shortest path between $q_k$ and $a$ over ConceptNet; \textit{min/max/mean} are computed over $\{L_1,L_2,L_3\}$ as three aspects to measure the overall difficulty.
Fig.\ (b), (c), and (d) show that generally \MyColorBox[FigRsBlue]{\textsc{RiddleSense}} has a longer question-answer path than \MyColorBox[FigCsqaOrange]{CommonsenseQA}, thus being harder to reason.
	}
	\label{fig:path} 
\end{figure*}


\section{Data Analysis of \textsc{RiddleSense}}
\label{sec:dataana}
In this section,
we first report the key statistics of the proposed \textsc{RiddleSense} dataset, 
then we compare it to CommonsenseQA~\cite{Talmor2018CommonsenseQAAQ} from two major angles: the distribution of the lengths of Q-A paths and the types of reasoning chains, which serve as an effective proxy to analyze the differences between the two datasets.

\begin{algorithm}[t]
    \SetAlgoLined
    \SetKwInput{KwRemarks}{Remarks}
    \SetKwFunction{FnExtract}{extractConcept}
    \SetKwFunction{FnShortestPath}{shortestPathLen}
    \SetKwFunction{FnDeg}{deg}
    \SetKwFunction{FnAppend}{append}
    \SetKwFunction{FnMin}{min}
    \SetKwFunction{FnMax}{max}
    \SetKwFunction{FnMean}{mean}
    \KwIn{Knowledge graph \(KG=(V,E)\), riddle question \(Q\), riddle answer \(A\)}
    \KwOut{\(minPathLength\), \(maxPathLength\), \(meanPathLength\)}
    
    \(QC \gets\) \FnExtract{\(Q\)}\\
    \(AC \gets\) \FnExtract{\(A\)}\\
    \(ac \gets v \in AC\) with smallest \FnDeg{\(G, v\)}\\
    \(l \gets\) []\\
     \ForEach{\(qc \in QC\)}{
        \(path \gets\) \FnShortestPath{KG, qc, ac}\\
        \If{\(path \neq None\)}{
            \(l\).\FnAppend{\(path\)}
        }
     }
    \(minPathLength \gets\) \FnMin(\(l\))\\
    \(maxPathLength \gets\) \FnMax(\(l\))\\
    \(meanPathLength \gets\) \FnMean(\(l\))
    \caption{Get statistics of QA paths.}
    \label{algo:path-len}
\end{algorithm}

\begin{table*}[th]
	\centering
	\scalebox{0.78
	}{
\begin{tabular}{c|c|c|c}
\toprule
 \multicolumn{4}{c}{\cellcolor{brown!25} \textbf{CommonsenseQA\quad(CSQA)}}                                                                                                                            \\ \midrule
\textbf{1-hop (\underline{14.0\%})}      & \textbf{2-hop (\underline{34.4\%})}                & \textbf{3-hop (\underline{41.5\%})}                           & \textbf{4-hop  (\underline{9.5\%})}                                      \\\midrule
AtLoc (4.8\%)  & \MyColorBox[red!15]{Related-Related }  (8.3\%)   & \MyColorBox[red!15]{Related-Related-Related }  (4.1\%)    & \MyColorBox[red!15]{Related $\times$ 4} ({0.4}\%)    \\
\MyColorBox[red!15]{Related} (3.4\%) & Related-AtLoc (4.5\%) & Related-Related-AtLoc (2.7\%) & Related $\times$ 3 -AtLoc (0.3\%) \\
Causes (1.1\%)      & Related-Antonym (1.8\%)     & Related-AtLoc$^{-1}$-AtLoc (1.4\%) & Related-Related-AtLoc$^{-1}$-AtLoc (0.3\%) \\
Antonym (0.9\%)     & Related-IsA$^{-1}$ (1.3\%)        & Related-Related-Antonym (1.3\%)      & Related $\times$ 3 -Antonym (0.2\%)      \\
Capableof (0.8\%)   & Related-AtLoc$^{-1}$ (0.9\%) & Related-Related-CapableOf (1.3\%)    & Related$\times$2-SubEvent$^{-1}$-Cause (0.1\%)    \\ 
... & ... & ... & ...  \\ \midrule
\rowcolor{yellow!20} $\rho=\frac{3.4}{4.8}=0.7$ & $\rho=\frac{8.3}{4.5}=1.8$  & $\rho=\frac{4.1}{2.7}=1.5$  & $\rho=\frac{0.4}{0.3}=1.3$   \\
\midrule

\midrule
\multicolumn{4}{c}{\cellcolor{cyan!15}\textbf{RiddleSense\quad(RS)}} \\ \midrule
\textbf{1-hop (\underline{4.6\%})}       & \textbf{2-hop (\underline{31.6\%}) }               & \textbf{3-hop (\underline{47.8\%}) }               & \textbf{4-hop (\underline{14.0\%}) }                                    \\ \midrule
\MyColorBox[red!15]{Related} (3.1\%)   & \MyColorBox[red!15]{Related-Related } (13.1\%)  & \MyColorBox[red!15]{Related-Related-Related }  (10.6\%)   &  \MyColorBox[red!15]{Related $\times$ 4} (1.8\%)    \\
Antonym ({0.4}\%)     & Related-Antonym (2.1\%)     & Related-Related-IsA$^{-1}$ (2.6\%)         & Antonym-Related$\times$3 (0.4\%)      \\
IsA$^{-1}$ (0.3\%)        & Related-IsA$^{-1}$ (2.0\%)        & Related-Related-Antonym (1.6\%)      & Related$\times$3 -IsA$^{-1}$ (0.3\%)         \\
PartOf (0.1\%)      & Related-AtLoc$^{-1}$ (1.3\%) & Related-Antonym-Related (1.5\%)      & Related$\times$2-IsA$^{-1}$-Related (0.3\%)         \\
AtLoc$^{-1}$ (0.1\%) & Antonym-Related (0.8\%)     & Antonym-{Related-Related } (1.5\%)      & Related$\times$2-Antonym-Related (0.3\%) \\ 
... & ... & ... & ...  \\
\midrule
\rowcolor{yellow!20} $\rho=\frac{3.1}{0.4}=7.8$ & $\rho=\frac{13.1}{2.1}=6.2$  & $\rho=\frac{10.6}{2.6}=4.1$  & $\rho=\frac{1.8}{0.4}=4.5$   \\
\bottomrule 
\end{tabular}
	} 
	
	\caption{The top-5 most frequent types of reasoning chains in CSQA and RS datasets, grouped by their length $k=\{1,2,3,4\}$.
	The \MyColorBox[yellow!20]{implicit-ratio $\rho$} is defined as the ratio of the implicit reasoning types (i.e., \MyColorBox[red!15]{Related$\times k$}) over the most frequent types with at least one explicit relation (e.g., AtLoc) of the same length $k$.
	}
	\label{tab:path}
\end{table*}

\subsection{Key Statistics}
Table~\ref{tab:stat} presents the key statistics of \textsc{RiddleSense} (RS) and the comparisons with CommonsenseQA (CSQA) which is the most similar benchmark to ours.
Although the size of \textsc{RS} is smaller than CSQA,
we argue that \textsc{RS} is complementary to the CSQA dataset and introduces novel challenges for the commonsense reasoning community.
As they share the same format, we can test different methods by training on either CSQA-only, RS-only, or the concatenation of CSQA and RS, as we show later in Section~\ref{sec:exp}.

Moreover, there is a greater number of long questions (i.e., containing more than 20 words) in RS than in CSQA.
Additionally, we find that RS questions have a lower normalized pseudo-likelihood (PLL)~\cite{salazar2019masked}, a proxy of estimating sentence probability, suggesting that RS questions are more puzzling (i.e., the words are less frequently co-occurring).
We also use a RoBERTa model fine-tuned on MNLI~\cite{Williams2018ABC} to perform natural language inference between CSQA/RS questions and their answers.
There is a much greater proportion of questions in RS that have \textit{conflicting} relations with their correct answers than compared to CSQA. 
This is indicative of RS's complexity due to the \textit{self-contradictory} and \textit{perplexing} nature of riddles.

Interestingly, we also find that although there are about twice as many examples in CSQA as RS, there are more distinct words in the questions and answer choices of RS than CSQA, suggesting that RS covers more diverse topics than CSQA.

\subsection{Distribution of the Lengths of Q-A Paths}
Our main intuition is that the shortest paths between question concepts and the answer concepts can approximate the \textit{underlying reasoning chains}, which are hidden and difficult to label.
To understand the difference between CSQA and RS in terms of their reasoning chains,
we use \textit{Q-A paths} over ConceptNet as a proxy. 
For a riddle question, a set of \textit{Q-A path lengths} are the lengths of the shortest paths between every question concept and the answer concept, i.e., \textit{shortestPathLen(KG, qc, ac)} in Alg. \ref{algo:path-len}.
For a question-answer pair, we first extract the concepts mentioned in the question and the answer respectively (\textit{extractConcept()} in Algorithm \ref{algo:path-len}), following the steps of~\citeauthor{kagnet-emnlp19} (\citeyear{kagnet-emnlp19}) and \citeauthor{feng2020scalable} (\citeyear{feng2020scalable}).
If there are three question concepts $\{q_1, q_2, q_3\}$ and an answer concept $a$, we denote their shortest path lengths as $\{L_1, L_2, L_3\}$.
Finally, we compute the min/max/mean over them for a comprehensive understanding of the approximated difficulty of this riddle 
--- a greater value indicates a more challenging example.

As shown in Figure~\ref{fig:path} (b), we can see that RS has longer Q-A paths as underlying reasoning chains.
In addition,
we can see that RS generally has longer chains, particularly the min of CSQA is $1$-hop for more than 80\% of examples.
On the other hand, only about 30\% of RS examples have 1-hop minimum Q-A paths, while about 50\% of the examples have 2-hop min Q-A paths.
The distribution over the maximum in Figure~\ref{fig:path} (d) also shows that RS tends to have longer maximum paths than CSQA.
We also show the percentage of all Q-A paths of different length as part of Table~\ref{tab:path},
and we can see that RS has longer paths in general (e.g., CSQA = 14.0\% vs. RS = 4.6\% in 1-hop).

\subsection{Relational Types of Reasoning Paths}
In addition to the analysis on path length, we also show that the relation types of Q-A paths for RS and CSQA have clear differences, as shown in Table~\ref{tab:path}.
The types of reasoning chains in RS rely more on a special relation in ConceptNet --- \texttt{Related}, which is relatively more implicit and can not be grounded to a specific, explicit relation such as \texttt{AtLoc} (e.g., \textless\textit{wind}, \texttt{Related}, \textit{air}\textgreater ~vs. \textless\textit{lamp}, \texttt{AtLoc}, \textit{table}\textgreater).
The most frequent relation between question concepts and answer concepts in CSQA is the \texttt{AtLoc} relation (4.8\%), however, it is \texttt{Related} (3.1\%) in RS.
We define \textit{implicit-ratio} for $k$-hop paths, 
$\rho_k=\frac{\% (\texttt{Related}\times k)}{\% (E_k)}$,
where $E_k$ is the most frequent type of chains with at least one explicit relation of length $k$.
In RS, $\rho_k$ is around $4.1\sim7.8$, while it is about $0.7\sim1.8$ for CSQA.
Thus, we conclude that the dominant reasoning chains in RS are much more implicit, and consequently RS is more challenging to reason with using commonsense knowledge resources like ConceptNet.

\begin{figure}[th!]
	\centering
	\includegraphics[width=1\linewidth]{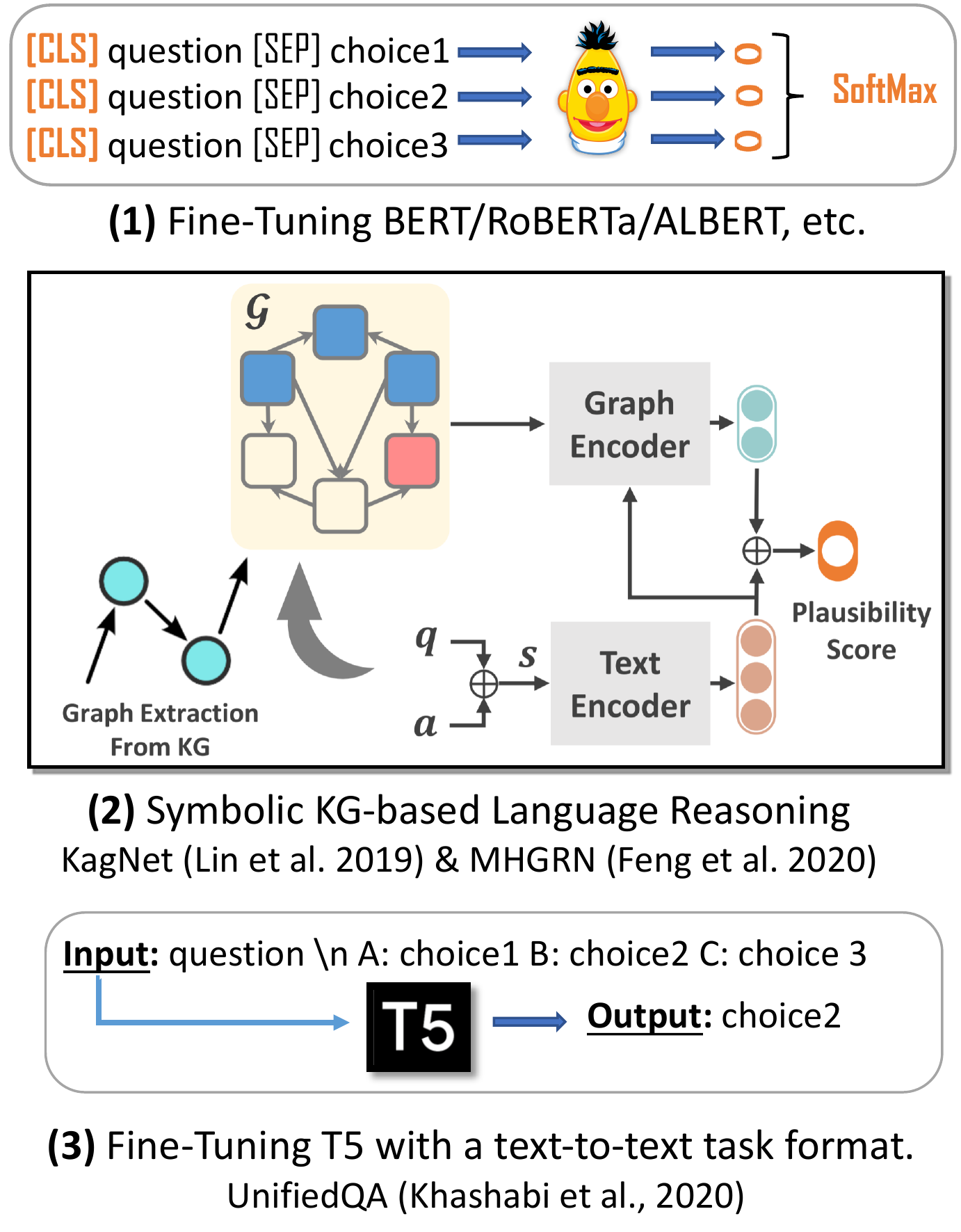}
	\caption{Three types of baseline methods: 1) fine-tuning pre-trained LMs, 2) incorporating graph-based reasoner, 3) fine-tuning a unified text-to-text LM. }
	\label{fig:models} 
\end{figure}

\section{Experiments}\label{sec:exp}
We first introduce three types of popular baseline methods for commonsense reasoning (Section~\ref{sec:baseline}), then we present our main experimental results with analysis (Section~\ref{sec:result}), and finally show case studies for error analysis (Section~\ref{sec:error}). 

\subsection{Baseline Methods}
\label{sec:baseline}
Given a riddle question $q$, there are 5 different choices $\{c_1, \dots, c_5\}$, where only one of them is the correct choice and the others are distractors.
The model needs to rank all choices and select the best one as the final answer.
There are three major types of models for commonsense reasoning tasks in this format: 
1) fine-tuning pretrained language models, 
2) incorporating relevant knowledge graphs for reasoning, 
3) fine-tuning a unified text-to-text QA model, 
as shown in Figure~\ref{fig:models}.

\paragraph{Fine-tuning Pre-trained LMs}
As we seek to investigate how well current NLU models can perform in higher-order commonsense reasoning,
we first experiment with a typical set of large pretrained language models such as BERT~\cite{Devlin2019BERTPO}, RoBERTa~\cite{Liu2019RoBERTaAR}, 
and ALBERT~\cite{Lan2020ALBERT}.
We concatenate the question with each choice, using \texttt{[SEP]} as the separator, thus forming a \textit{statement}.
Then, we fine-tune any pretrained LMs like BERT to use their \texttt{[CLS]}  token embeddings to predict a score for each statement.
Then, a set of five scores about an example will be fed to SoftMax to optimize for maximizing the score of the correct choice.

\paragraph{LMs + Graph Reasoning Modules}
KagNet~\cite{kagnet-emnlp19} and MHGRN~\cite{feng2020scalable} are two typical graph-based language reasoning models.
They both extract a \textit{schema graph} from ConceptNet, i.e., a subgraph of ConceptNet consisting of Q-A paths in Figure~\ref{fig:path},
by incorporating them with a graph encoding module.
They finally fuse the external commonsense knowledge with a text encoder (e.g., a pretrained LM). 
KagNet uses heuristics to prune irrelevant paths and then encode them with path-based LSTM and hierarchical attention to select the most important paths for improving commonsense reasoning.
In contrast, the recent MHGRN explicitly encodes multi-hop paths at scale using graph networks with relational attention, improving efficiency and performance over KagNet and other models.
A unique merit of such graph-based models is their \textit{interpretibility} due to the neural attention over the symbolic structures of KGs.

\paragraph{Fine-Tuning a Text-to-Text QA Model}
{UnifiedQA}~\cite{khashabi2020unifiedqa}, the state-of-the-art multiple-choice QA model, simply concatenates the question with all answer candidates as a single input sequence to a T5~\cite{t5} model for learning to generate the correct choice as extracting a span from the input.
Apart from the multiple-choice QA format, it is also trained with other QA task formats so that it can benefit from many other QA datasets (including CSQA) via sharing the model parameters.

\paragraph{Human Evaluation}
We invite three native English speakers who study computer science to solve 100 riddle examples sampled from the test set.
They achieved an average accuracy of 91.3\%.


\begin{table*}[th!]
	\centering
	\scalebox{0.92
	}{
		\begin{tabular}{c||cc||cc||cc}
		\toprule
              \multicolumn{1}{c||}{Models $\downarrow$ Training Data $\rightarrow$}& \multicolumn{2}{c||}{\cellcolor{brown!25}Train = CSQA} & \multicolumn{2}{c||}{\cellcolor{cyan!20}Train = RiddleSense} & \multicolumn{2}{c}{\cellcolor{blue!15}Train = RS+CSQA} \\ \midrule
            RiddleSense-Split   $\rightarrow$  & Dev          & Test        & Dev         & Test       & Dev            & Test           \\ \midrule
\rowcolor{gray!20} \textit{Random Guess} &  20.0     &  20.0   & 20.0      &  20.0       &     20.0            &  20.0  \\ \midrule
BERT-Base~\cite{Devlin2019}    & 33.59         & 34.61        & 54.16        & 42.43       & 56.22           & 47.67           \\
BERT-Large~\cite{Devlin2019}     & 36.14         & 39.10        & 55.24        & 45.09       & 57.69           & 54.91           \\
RoBERTa-Large~\cite{Liu2019RoBERTaAR}  & 43.68         & 47.42        & 60.72        & 52.58       & 66.11           & 59.82           \\
ALBERT-XXL~\cite{Lan2020ALBERT} & \textbf{51.03}         & \textbf{51.00 }       & \underline{66.99}        & \underline{60.65}       &\textbf{ 71.50 }          &\underline{ 67.30  }         \\ 
\midrule
KagNet (RoBERTa-L)~\cite{kagnet-emnlp19}  &   42.66  & 48.24 & 61.77  & 53.72  & 66.55  & 59.72   \\
MHGRN (RoBERTa-L)~\cite{feng2020scalable} & 46.83 & 49.65 & 63.27 & 54.49 & 66.90 & 63.73       \\
MHGRN (ALBERT-XXL)~\cite{feng2020scalable}  &  \underline{50.89}         & {50.21}       & {66.27}        & {59.93}       & \underline{70.81}          &{ 66.81  }         \\ \midrule
UnifiedQA (T5-Large)~\cite{khashabi2020unifiedqa}   &  28.50 & 37.27 & 56.21 & 56.40 & 58.17 & 56.57 \\
UnifiedQA (T5-3B)~\cite{khashabi2020unifiedqa}   & 37.32         & \underline{50.25}       & \textbf{67.38}        & \textbf{66.06}       &        68.26              &       	\textbf{68.80} \\ \midrule
\rowcolor{gray!20} \textit{Human Performance}  &  -     &  91.33   & -      &  91.33       &      -            &   91.33  \\ \bottomrule
\end{tabular}
	} 
	
	\caption{Benchmark performance over the dev and test set of \textsc{RiddleSense}  .}
	\label{tab:results}
\end{table*}

\begin{figure}[t]
	\centering 
	\includegraphics[width=1\linewidth]{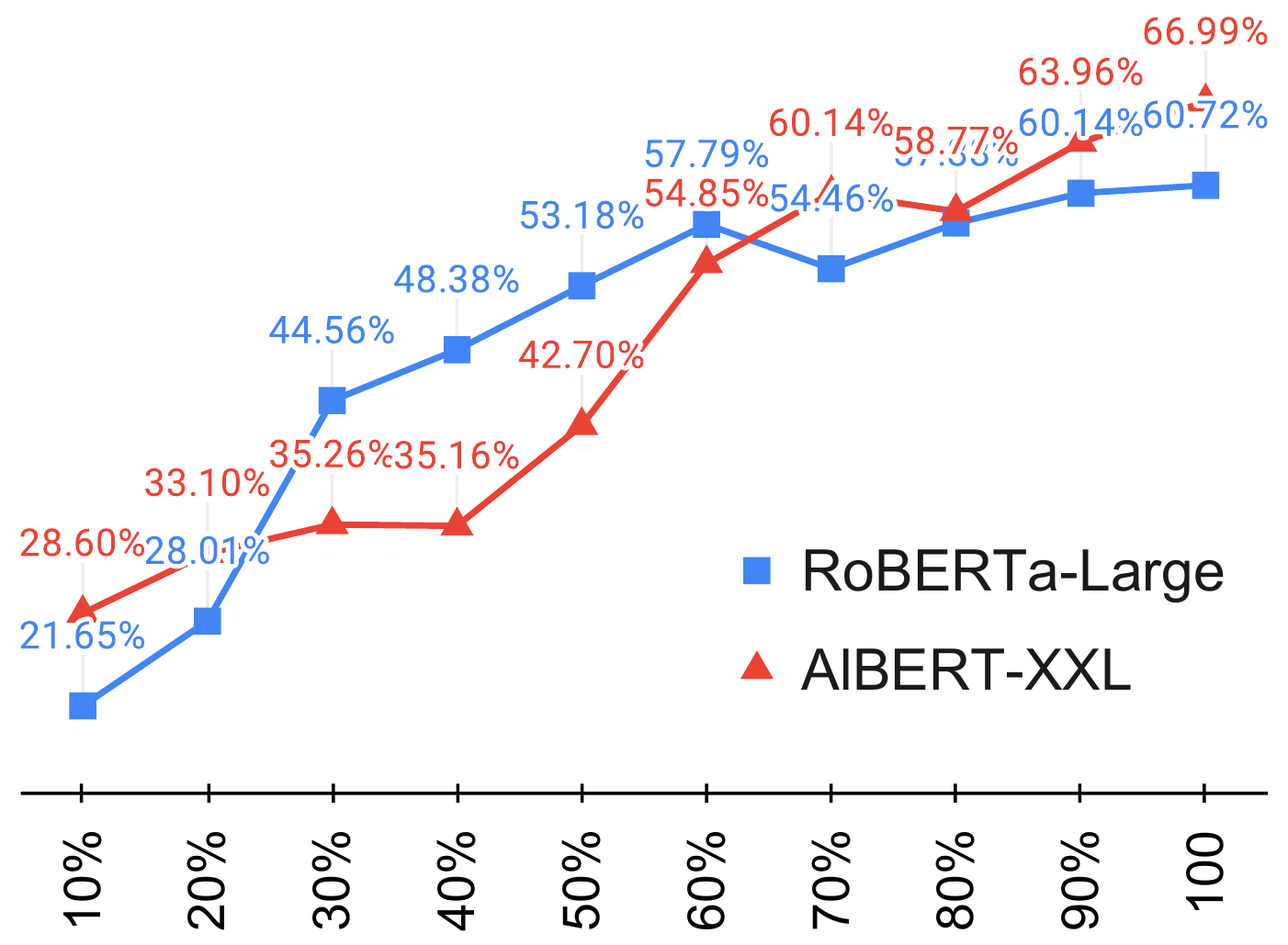}
	\caption{The curve of dev accuracy using different percentage of the RS-training data, respectively for RoBERTa-Large and ALBERT-XXL. }
	\label{fig:curve} 
\end{figure}

\subsection{Results and Analysis}
\label{sec:result}

We show the main results of the experiments in Table~\ref{tab:results}.
There are 3 settings according to the different training data options: 1) the training data of CSQA, 2) the training data of RS, and 3) the concatenation of both RS and CSQA, while all experiments are validated over the dev set of RS.
However, as the public UnifiedQA checkpoints were already trained on CSQA (together with many other QA datasets), we directly use them for inference over RS in the first setting (i.e., ``Train=CSQA'').
This also suggests that the performance of UnifiedQA models in 2nd setting should be better than others although they all are fine-tuned on RS's training data only.

We can see that larger pretrained language understanding models always gain better performance, ranging from BERT-base to {Albert-XXL}, which gets the best performance in this group of baselines (67.30\%).
This matches their performance comparisions on CSQA and other benchmark datasets as well, suggesting that a better pre-trained language model can be also identified by \textsc{RiddleSense} as well.
Interestingly, we find that ALBERT-XXL is so powerful that it can generalize from training on CSQA only but achieve comparable results with RoBERTa-Large that is trained over RS (i.e., 51.0\% vs. 52.6\%).
However, if we look at the curve of dev accuracy when using different percentage of the RS-train data (setting 2) in Figure~\ref{fig:curve}, 
we can see that RoBERTa-Large can generally outperform ALBERTA-XXL when using less than 60\% data for fine-tuning.



Moreover, we find that the {KG-enhanced models}, KagNet and MHGRN, 
using RoBERTa-Large (RB-L) as the encoder, perform better than vanilla RB-L.
Although the Q-A paths over ConceptNet have more implicit paths (e.g., \texttt{Related}$\times k$),
some paths can still be beneficial. For example, $$\textit{wind} \xleftrightarrow[]{~\texttt{Related}~} \textit{blow} \xleftrightarrow[]{~\texttt{Related}~} \textit{candle},$$
can still help reason about the riddle \textit{``... Wind is my foe. What am I?''} to the answer ``candle.''

The fusion of ConceptNet also improves in the situation when only training with CSQA data using RoBERTa-Large.
However, the improvement of KagNet is negative, which is unexpected. 
We conjecture that this is because the extracted subgraphs from the ConceptNet does not guarantee the reasoning path from question concepts to answer concepts, while the training phase \textit{forces} models to learn to reason over those graphs, yielding a possibly \textit{harmful} impact.
Additionally, we find that MHGRN with ALBERT-XXL also results in a worse performance, unlike using RoBERTa-Large.
We believe this may be related to the specific design of ALBERT, which reuses model parameters for multiple layers, and thus it could be a problem when fused with another learnable module (e.g., a graph network in MHGRN).

Fine-tuning UnifiedQA with T5-3B achieves the best performance, which is also the case for CSQA in their leaderboard. 
This is expected for two reasons: 1) UnifiedQA has been trained over multiple other QA datasets, which increases its generalization ability, 2) UnifiedQA considers all choices together at a time and thus can better compare different choices with self-attention mechanism of Transformer~\cite{vaswani2017attention}.



\begin{figure*}[th!]
	\centering
	\includegraphics[width=1\linewidth]{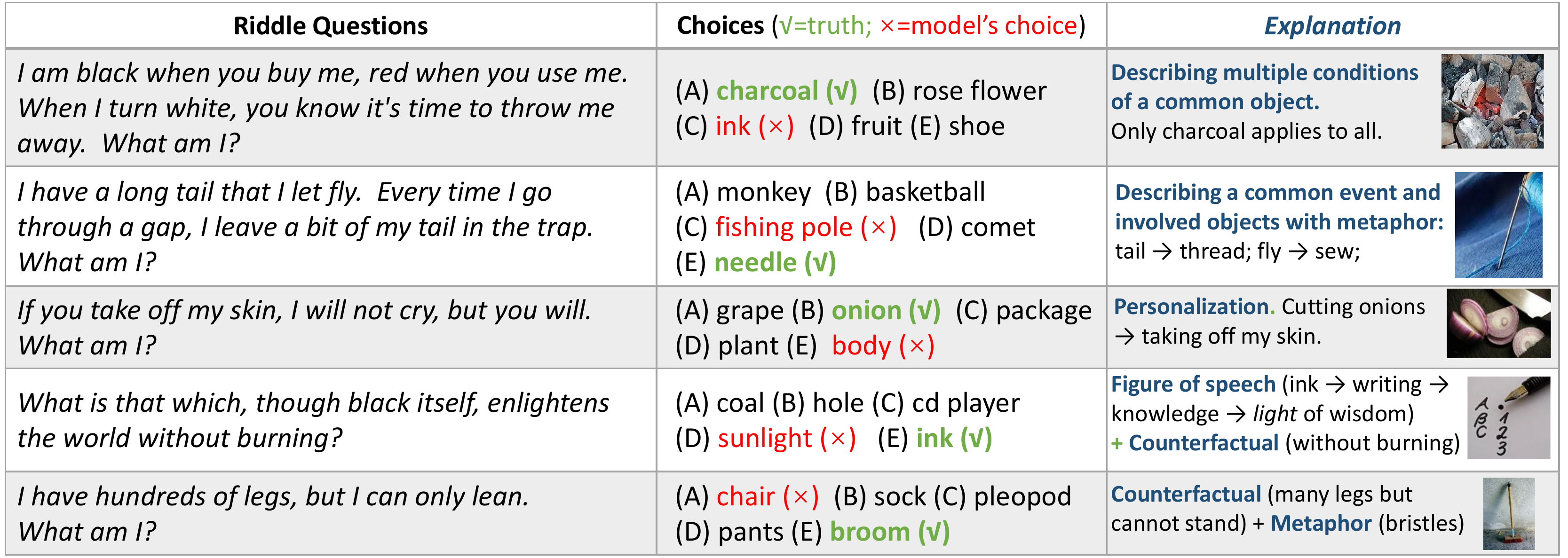}
	\caption{Case studies of the error by UnifiedQA-3B model on the test set of \textsc{RiddleSense}.}
	\label{fig:error} 
\end{figure*}

\subsection{Error Analysis and Future Directions}
\label{sec:error}
We show a few examples that are mistakenly predicted by the UnifiedQA-3B model in Figure~\ref{fig:error}.
From these concrete cases, 
we can see that even the best model cannot solve riddles that can be trivial to humans, especially when there are metaphors and/or counterfactual situations. 
We argue that future research should aim to address the creative use of language in commonsense reasoning and general understanding of language, 
as creativity is a critical feature of natural language.
We list several promising directions as follows.

First of all, we should \textit{mine (semi-)structured knowledge of metaphors}, so that concepts can connect via metaphorical links (e.g., ``tail'' $\rightarrow$ ``thread'').
Second, to prevent false inferences, we need \textit{more complete, precise commonsense knowledge of concepts}. For example, in Figure~\ref{fig:error}, a model should know a chair only has exactly \textit{four} legs instead of \textit{hundreds}~\cite{lin2020birds}; ink can be black or red, but it won’t change over time. However, current KGs only have (leg, PartOf, chair) and (ink, HasProperty, black/red). 
In addition, the reasoning methods should incorporate more \textit{symbolic logic rules}, so that the multi-hop conditions and counterfactual ``but-no'' negations will be handled better. 
Finally, we think the graph-augmented methods should be improved to \textit{compare multiple options} in a schema graph, e.g., \textsc{QA-GNN}~\cite{yasunaga2021qagnn}.  
Both \textsc{KagNet} and \textsc{MHGRN} consider only a single option at a time which prevents them from effectively reasoning about the subtle differences between options.

\section{Related Work}\label{sec:rel_work}

\subsection*{Benchmarking Machine Common Sense}
 
The prior works on building commonsense reasoning benchmarks touch different aspects of commonsense reasoning:
SWAG~\cite{Zellers2018SWAGAL}, HellaSWAG~\cite{Zellers2019HellaSwagCA}, CODAH~\cite{Chen2019CODAHAA}, aNLI~\cite{bhagavatula2019abductive} for situation-based reasoning;
Physical IQA~\cite{bisk2020piqa} on physical knowledge;
Social IQA~\cite{sap-etal-2019-social} on social psychology knowledge;
LocatedNearRE~\cite{Xu2018AutomaticEO} on mining spatial commonsense knowledge;
DoQ~\cite{elazar2019large} and NumerSense~\cite{lin2020birds} on numerical common sense;
CommonGen~\cite{lin2019commongen} for generative commonsense reasoning, and many others;
OpenCSR~\cite{lin2021opencsr} and ProtoQA~\cite{Boratko2020ProtoQAAQ} aim to test commonsense reasoning ability in an open-ended setting.

CommonsenseQA~\cite{Talmor2018CommonsenseQAAQ} has the same format as our proposed \textsc{RiddleSense}, and both target general commonsense knowledge via multiple-choice question answering.
However, CSQA focuses more on straightforward questions where the description of the answer concept is easy to understand and retrieval over ConceptNet, while RS makes use of riddle questions to test higher-order commonsense reasoning ability.
More detailed comparisions between them are in Section~\ref{sec:dataana}, which shows that the unique challenges of the RiddleSense on multiple dimensions.

\subsection*{Commonsense Reasoning Methods}
Our experiments cover three major types of commonsense reasoning methods that are popular in many benchmarks: fine-tuning pretrained LMs~\cite{Devlin2019, Liu2019RoBERTaAR, Lan2020ALBERT}, graph-based reasoning with external KGs~\cite{kagnet-emnlp19, feng2020scalable}, and fine-tuning unified text-to-text QA models~\cite{khashabi2020unifiedqa}.
Apart from ConceptNet, 
There are also some methods~\cite{lv2019graph, xu2020fusing}
using additional knowledge resources such as Wikipedia and Wiktionary.
A few recent methods also aim to generate relevant triples via language generation models so that the context graph is more beneficial for reasoning~\cite{wang2020connecting, yan2020learning}.
Our experiments in this paper aim to compare the most typical and popular methods which have open-source implementations,
which we believe are beneficial for understanding the limitation of these methods in higher-order commonsense reasoning --- \textsc{RiddleSense}.

\subsection*{Computational Creativity and NLP}
Creativity has been seen as a central property of the human use of natural language~\cite{mcdonald1994creative}.
Text should not be always taken at face value, however, higher-order use of language and figurative devices such as metaphor can communicate richer meanings and needs deeper reading and more complicated reasoning skills~\cite{veale2011creative}.
Recent works on processing language with creative use focus on metaphor detection~\cite{gao2018neural}, pun generation~\cite{He2019PunGW, Luo2019PunGANGA}, creative story generation, and humor detection~\cite{Weller2019HumorDA, Weller2020TheRD}, sarcasm generation~\cite{chakrabarty2020r}, etc. 

Riddling, as a way to use creative descriptions to query a common concept, are relatively underexplored.
Previous works~\cite{tan2016solving, oliveira2018exploring} focus on the generation of riddles in specific languages and usually rely on language-specific features (e.g., decomposing a Chinese character into multiple smaller pieces).
There is few datasets or public resources for studying riddles as a reasoning task, to the best of our knowledge. 
The proposed \textsc{RiddleSense} is among the very first works connecting commonsense reasoning and computational creative, and provides a large dataset to train and evaluate models for answering riddle questions.


\section{Conclusion}\label{sec:conclusion}
 We propose a novel commonsense reasoning challenge, \textsc{RiddleSense}, which requires complex commonsense skills for reasoning about creative and counterfactual questions, coming with a large multiple-choice QA dataset.  
 We systematically evaluate recent commonsense reasoning methods over the proposed \textsc{RiddleSense} dataset, and find that the best model is still far behind human performance, suggesting that there is still much space for commonsense reasoning methods to improve.
 We hope \textsc{RiddleSense} can serve as a benchmark dataset for future research targeting complex commonsense reasoning and computational creativity.

\section*{Acknowledgements}
This research is supported in part by the Office of the Director of National Intelligence (ODNI), Intelligence Advanced Research Projects Activity (IARPA), via Contract No. 2019-19051600007, the DARPA MCS program under Contract No. N660011924033 with the United States Office Of Naval Research, the Defense Advanced Research Projects Agency with award W911NF-19-20271, and NSF SMA 18-29268. The views and conclusions contained herein are those of the authors and should not be interpreted as necessarily representing the official policies, either expressed or implied, of ODNI, IARPA, or the U.S. Government. We would like to thank all the collaborators in USC INK research lab and the reviewers for their constructive feedback on the work.

\section*{{Ethical Considerations}}

\paragraph{Copyright of Riddles.} 
The RiddleSense dataset is consistent with the terms of use of the fan websites and the intellectual property and privacy rights of the original sources.
All of our riddles and answers are from fan websites that can be accessed freely.
The website owners state that we may print and download material from the sites solely for \textit{non-commercial use} provided that we agree not to change or delete any copyright or proprietary notices from the materials.
Therefore, 
in addition to the dataset itself, we also provide the according copyright statements of every website and an URL link to the original page for each riddle. 
The dataset users must sign an informed consent form that they will only use our dataset for \textit{research purposes} before they can access the both the riddles and our annotations.

\bibliography{riddleqa_rebiber} 
\bibliographystyle{acl_natbib}

\end{document}